%% file: main.tex
\documentclass[10pt,twocolumn,letterpaper]{article}

\usepackage[pagenumbers]{iccv} % To force page numbers, e.g. for an arXiv version
\usepackage{algorithm}
\usepackage{algpseudocode}
\usepackage{graphicx}
\usepackage{array} 
\usepackage{float}
\usepackage{caption}


\definecolor{iccvblue}{rgb}{0.21,0.49,0.74}
\usepackage[pagebackref,breaklinks,colorlinks,allcolors=iccvblue]{hyperref}

\def\paperID{11393} % *** Enter the Paper ID here
\def\confName{ICCV}
\def\confYear{2025}

\title{InstructAttribute: Fine-grained Object Attributes editing with Instruction}

\author{
    Xingxi Yin \quad Jingfeng Zhang  \quad Yue Deng \quad Zhi Li \quad Yicheng Li \quad Yin Zhang\\
    Zhejiang University\\
    Hangzhou, China\\
    {\tt\small \{12321103\}@zju.edu.cn}
}

\begin{document}

\maketitle
\maketitle

\input{sec/0_abstract}    
\input{sec/1_intro}
\input{sec/2_relatedWork}
\input{sec/3.preliminaries}  
\input{sec/3_data_engine}

\input{sec/4_InstructionAttribute}
\input{sec/5_experiment}
\input{sec/6_conclusion}

{
    \small
    \bibliographystyle{ieeenat_fullname}
    \bibliography{main}
}

\end{document}

%% file: sec/0_abstract.tex
\begin{abstract}
Text-to-image (T2I) diffusion models are widely used in image editing due to their powerful generative capabilities. However, achieving fine-grained control over specific object attributes, such as color and material, remains a considerable challenge. Existing methods often fail to accurately modify these attributes or compromise structural integrity and overall image consistency. To fill this gap, we introduce Structure Preservation and Attribute Amplification (SPAA), a novel training-free framework that enables precise generation of color and material attributes for the same object by intelligently manipulating self-attention maps and cross-attention values within diffusion models. Building on SPAA, we integrate multi-modal large language models (MLLMs) to automate data curation and instruction generation. Leveraging this object attribute data collection engine, we construct the Attribute Dataset, encompassing a comprehensive range of colors and materials across diverse object categories. Using this generated dataset, we propose InstructAttribute, an instruction-tuned model that enables fine-grained and object-level attribute editing through natural language prompts. This capability holds significant practical implications for diverse fields, from accelerating product design and e-commerce visualization to enhancing virtual try-on experiences. Extensive experiments demonstrate that InstructAttribute outperforms existing instruction-based baselines, achieving a superior balance between attribute modification accuracy and structural preservation.

\end{abstract}

%% file: sec/1_intro.tex
\section{Introduction}
\label{sec:intro}

In the burgeoning era of artificial intelligence-generated content (AIGC), diffusion probabilistic models (DPMs) \cite{ho2020denoising,song2020score} have fundamentally reshaped image generation, surpassing Generative Adversarial Networks (GANs) \cite{choi2018stargan, choi2020stargan, goodfellow2014generative, karras2019style, brock2018large} with their remarkable ability to synthesize highly realistic images across diverse scenarios. This tremendous progress has propelled diffusion-based Text-to-Image (T2I) models \cite{ho2022classifier, nichol2021glide, saharia2022photorealistic,rombach2022high, ramesh2022hierarchical, meng2023distillation,gu2022vector, podell2023sdxl} to widespread adoption for various image editing tasks. Current methods have demonstrated abilities to modify spatial arrangements \cite{epstein2023diffusion,chen2023anydoor}, adjust size and shape \cite{patashnik2023localizing, epstein2023diffusion, li2023layerdiffusion, goel2023pair}, alter actions and poses \cite{li2023layerdiffusion, kawar2023imagic, cao2023masactrl, huang2023kv}, colorize grayscale images \cite{weng2023cad, huang2022unicolor,liang2024control}, extract materials \cite{lopes2024material}, or manipulate perspectives \cite{cao2023masactrl}. Despite these advances, a critical gap remains: the precise, fine-grained control over specific visual attributes, such as the color and material of individual objects within an image, has not yet been fully achieved. Existing methods \cite{mokady2023null, hertz2022prompt, liu2024towards, brooks2023instructpix2pix, geng2024instructdiffusion, zhang2024hive, Zhang2023MagicBrush}, while showing promise in general image modification, frequently struggle to accurately alter these attributes without compromising the object's structural integrity or the consistency of other regions in the image.

This limitation significantly impacts real-world applications, particularly within the design and visualization industries. Imagine the inefficiencies in product design, where artists could instantly prototype new product variations, a car in a different paint color, a piece of furniture with a new fabric, without costly physical mock-ups. Consider the transformative potential for e-commerce visualization, allowing customers to dynamically explore clothing, home decor, or automotive options in a myriad of colors and materials, enhancing virtual try-on experiences and accelerating purchasing decisions. It is this precise, on-demand object attribute manipulation that our work aims to address.

\begin{figure}[!h]
    \centering
    \includegraphics[width=0.5\textwidth]{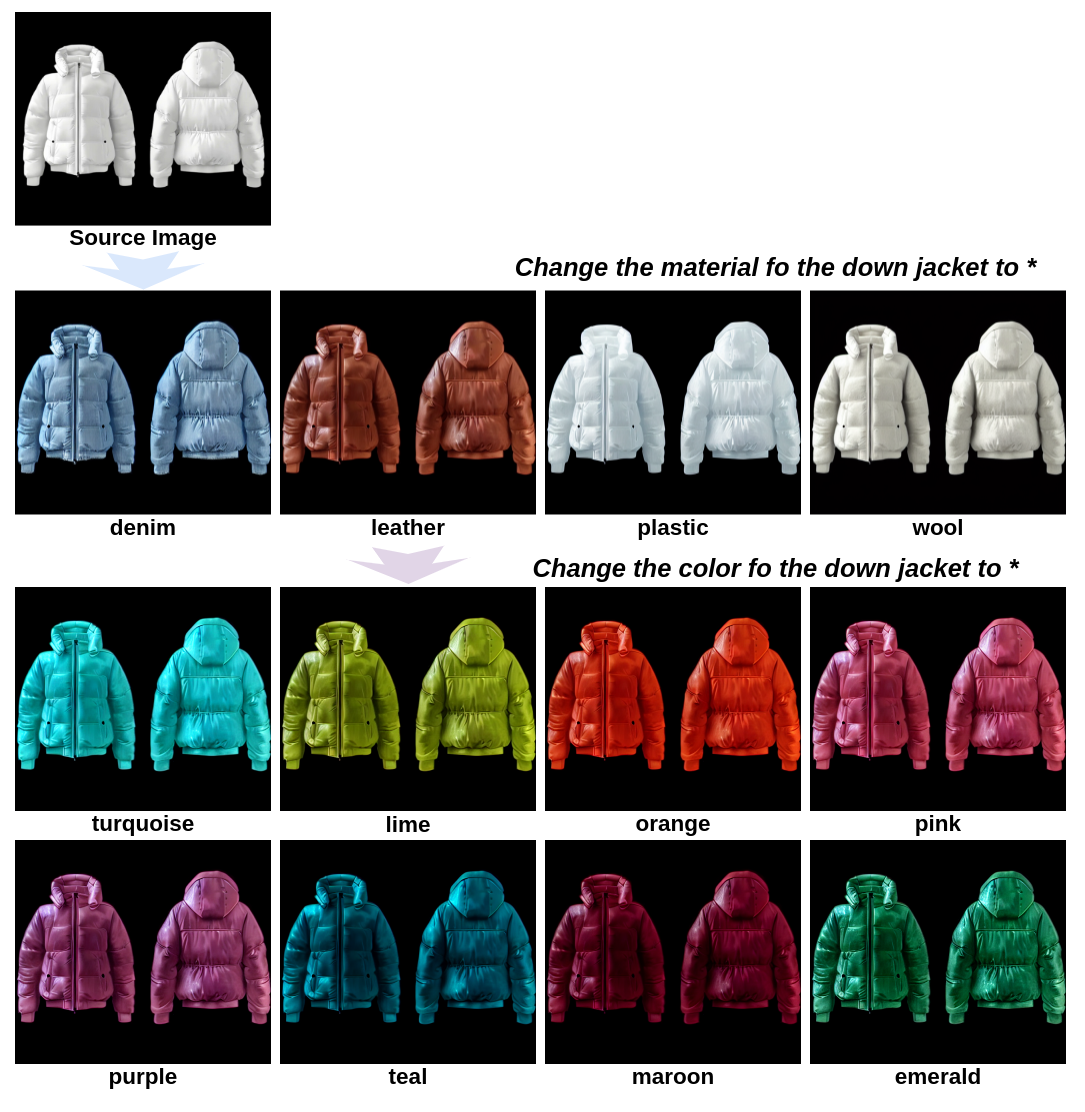}

    \caption{Given an unrendered 3D down jacket image with front and back views, our InstructAttribute enables precise, fine-grained editing of object attributes through instruction. The material of the down jacket is initially transformed to denim, leather, plastic and  wood. Then, the color of the leather jacket is further refined to turquoise, lime, orange, pink, purple, teal, maroon and emerald. See the consistency of the material and color change across the front and back views of the down jacket.}
    \label{fig:material_color_change}
\end{figure}

In this paper, we focus on tackling this challenge of object-level attribute alteration, specifically the manipulation of color and material, enabling users to modify these characteristics through intuitive textual instructions (see an example in Fig.~\ref{fig:material_color_change}). To achieve this goal, we propose a simple yet effective training-free approach, called \textbf{Structure Preservation and Attribute Amplification (SPAA)}, to generate high-quality fine-grained synthetic images with diverse color and material variations for the same object.  Then we introduce a data collection engine that utilizes advanced multimodal large language models, LLaVa \cite{liu2023llava}, to remove inappropriate target images. Leveraging the in-context learning capabilities of GPT-4o \cite{hurst2024gpt}, we generate a diverse set of instruction templates or object attribute modification and construct a comprehensive \textbf{Object-Attribute dataset}. Furthermore, based on the dataset generated with our engine, we develop an instruction-tuned model, \textbf{InstructAttribute}, which empowers users to execute precise color and material editing tasks using natural language commands.

In summary, our contributions are threefold.

    1) We investigated the roles of self-attention maps in preserving object structure and cross-attention's key and value matrices in influencing object attributes. Based on these insights, we developed Structure Preservation and Attribute Amplification (SPAA), a straightforward yet powerful method for synthesizing objects with varied visual characteristics, including color and material.

    2) We introduce the first high-quality, fine-grained object attribute dataset specifically designed for object color and material editing, named the Attribute Dataset. It consists of more than 3,300 subjects, 43 different colors, 14 different materials, and more than 1.1 million pairs of image pairs of source-target attributes.  

    3) We present InstructAttribute, an instruction-based method that uses natural language prompts to edit object attributes, specifically targeting color and material modifications.    
Comprehensive experiments demonstrate that our method achieves an optimal balance between effective color and material modification and superior preservation of structural and background details, outperforming existing state-of-the-art techniques.

\begin{figure*}[!ht]
    \centering
    \includegraphics[width=1.0\textwidth]{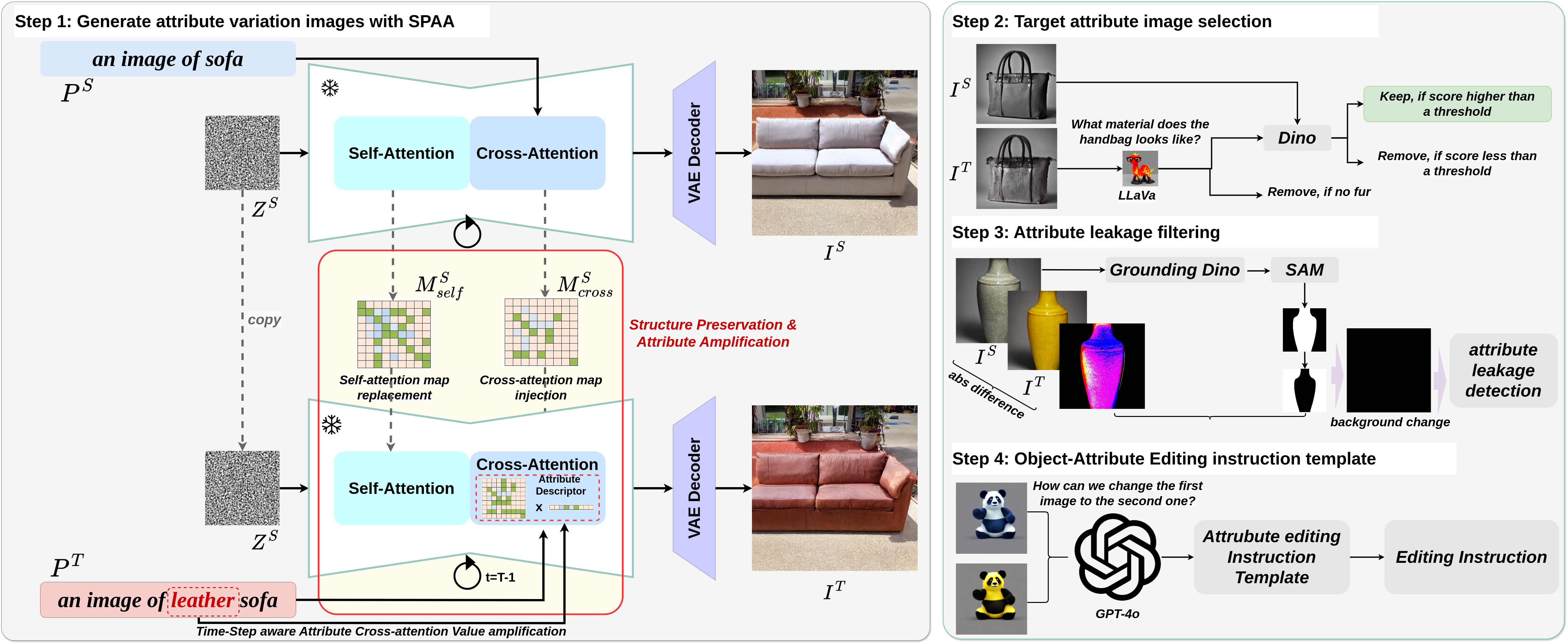}

    \caption{Pipeline of the Attribute Dataset Collection Engine. This figure illustrates our comprehensive pipeline for constructing a robust attribute dataset. First, we employ our proposed Structure Preservation and Attribute Amplification (SPAA) method to synthetically generate images of the same object while varying its attributes. Next, in a scalable data collection framework, we identify target attribute images (Step 2) and filter out instances of attribute leakage into the background (Step 3). Finally, we construct a diverse set of attribute editing instruction, completing the dataset for precise attribute manipulation.} 
    \label{fig:pipeline}
\end{figure*}

%% file: sec/2_relatedWork.tex
\section{Related work}
\label{sec:related_work}

A. Text-Guided Image Editing

Recent advances in diffusion-based text-guided image generation techniques have paved the way for the development of efficient text-based image editing approaches. Text-guided image editing methods modify images using textual descriptions, providing high-level or descriptive guidance to align the output with the specified text. Among these approaches, SDEdit \cite{meng2021sdedit} employs a two-step process of noise addition followed by denoising to align the output with the given prompts. Imagic \cite{kawar2023imagic} optimizes a text embedding aligned with the input image, interpolating it with the target description to produce contextually modified images. Prompt-to-Prompt (P2P) \cite{hertz2022prompt} leverages cross-attention layers to capture interactions between pixel structures and prompt words, allowing semantic editing through cross-attention map manipulation.  To extend P2P's \cite{hertz2022prompt}  capability to real images, Null Text Inversion (NTI) \cite{mokady2023null} was introduced, which updates null text embeddings for precise reconstruction while accommodating classifier-free guidance. EDICT \cite{wallace2023edict} proposes an image inversion approach using two noise vectors, enhancing both image reconstruction and textual fidelity. DiffEdit \cite{couairon2022diffedit} automatically generates an editing mask by comparing the source and target text prompts, pinpointing areas that need changes while preserving unchanged regions.  Both PnP \cite{tumanyan2023plug}, MasaCtrl \cite{cao2023masactrl} and FreePromptEditing (Free) \cite{liu2024towards} 
investigate the utilization of self-attention mechanisms and feature injection techniques to enhance image editing performance.  Both L-CAD \cite{weng2023cad}, UniColor \cite{huang2022unicolor} and Control Color \cite{liang2024control} focused on Image Colorization, specifically addressing the challenge of language-based or multimodal image colorization.  The material palette \cite{lopes2024material} automatically identifies and quantifyes the distinct materials present in a photograph, and extracts their physically-based rendering (PBR) properties. Text-guided image editing, while powerful, often struggles with semantic ambiguity \cite{elsharif2025visualizing, li2024text},  lack of fine-grained control \cite{wang2025training}, and potential for inconsistencies \cite{li2024text,meiri2023fixed}, as it relies on models interpreting broad descriptive prompts.

B. Instruction-Based Image Editing

Unlike text-guided image editing, instruction-based image editing focuses on interpreting and executing explicit task-oriented commands, allowing detailed edits without the need for extensive prompt engineering. InstructPix2Pix (IP2P) \cite{brooks2023instructpix2pix} is the pioneering study that enables image editing based on human instructions. Trained on a synthetic instruction-based image editing dataset which is generated with P2P \cite{hertz2022prompt} and GPT tools, IP2P facilitates image editing following user-provided instructions.  FoI \cite{guo2023focus} uses the implicit grounding capabilities of IP2P \cite{brooks2023instructpix2pix} to identify and refine specific editing regions, while employing cross-condition attention modulation to align each instruction with its target area, thereby reducing interference across multiple instructions. Later works, such as HIVE \cite{zhang2024hive} and MagicBrush \cite{Zhang2023MagicBrush}, focus on improving the quality or quantity of the dataset. HIVE \cite{zhang2024hive} introduces more training triplets and incorporates Reinforcement Learning from Human Feedback (RLHF) into instructional image editing to enhance alignment between edited images and human instructions.  MagicBrush \cite{Zhang2023MagicBrush} created a manually annotated instruction-guided image editing dataset by engaging human participants to perform edits using an online image editing tool. InstructDiffusion \cite{geng2024instructdiffusion} serves as a unified framework that conceptualizes a wide range of vision tasks as intuitive, human-like image manipulation processes. Emu Edit \cite{Sheynin2023EmuEP} is another multitask image editing model based on instruction, trained on a unified dataset that includes both image editing and recognition tasks. MGIE \cite{fu2023guiding} integrates an MLLM, LLaVa \cite{liu2023llava}, with Stable Diffusion to enable more precise and context-sensitive image editing. Similarly, SmartEdit \cite{huang2024smartedit} leverages LLaVA and incorporates a Bidirectional Interaction Module (BIM) to enable effective image editing in complex scenarios.  HQ-Edit \cite{hui2024hq} employs a novel, scalable pipeline that leverages advanced foundation models, GPT-4V \cite{openai2023gpt4} and DALL·E 3 \cite{openai2023dalle3}, to generate high-quality input-output image pairs accompanied by detailed textual editing instructions.  UltraEdit \cite{zhao2024ultraeditinstructionbasedfinegrainedimage} uses LLMs combined with in-context examples from human annotators to generate a diverse set of editing instructions, thereby enhancing the richness and variety of the dataset and addressing the previously unmet need for region-specific editing capabilities. Unlike those instruction-based methods \cite{brooks2023instructpix2pix,geng2024instructdiffusion,fu2023guiding,Zhang2023MagicBrush,huang2024smartedit,hui2024hq,zhao2024ultraeditinstructionbasedfinegrainedimage} that address a diverse array of image editing tasks, our InstructAttribute focuses exclusively on the color and material of an object, achieving fine-grained editing of these attributes.

%% file: sec/3.preliminaries.tex
\section{Preliminaries}
\label{sec:prelimiaries}

\textbf{Attention Mechanism.} In diffusion model, the attention mechanism \cite{ashish2017attention} plays a crucial role in aligning textual descriptions with visual elements to generate high-quality, coherent images. And the attention mechanism can be formulated as follows:
\[Attention(Q, K, V) =softmax\left(\frac{QK^{T}}{\sqrt{d_k}}\right)V, \quad (1) \]
Here, $Q$ represents the Query matrices, while $K$ and $V$ denote the Key and Value matrices. Within the U-Net \cite{ronneberger2015u}, self-attention helps the model to understand spatial relationships between different parts of the image, and self-attention maps can preserve spatial and structural information of an image \cite{tumanyan2023plug, cao2023masactrl, liu2024towards}. For cross-attention, query $Q$, key $K$, and value $V$ are obtained through the learned linear projection function $f_{Q}$, $f_{K}$, and $f_{V}$ separately.
\[
Q = {f_{Q}}(\phi(z_t)),  K = {f_{K}} (\phi(P)), V = {f_{V}} (\phi(P)), \quad (2) 
\]
It is obvious that the textual information encoding $\phi(P)$ is fed into the linear projection function $f_{K}$, $f_{V}$, which means that the text prompt affected the image synthesis process through the Key and Value matrices.

%% file: sec/3_data_engine.tex
\section{Object-Attribute Data Engine}
\label{sec:data_engine}

\begin{figure}[!ht]
    \centering
    \includegraphics[width=0.5\textwidth]{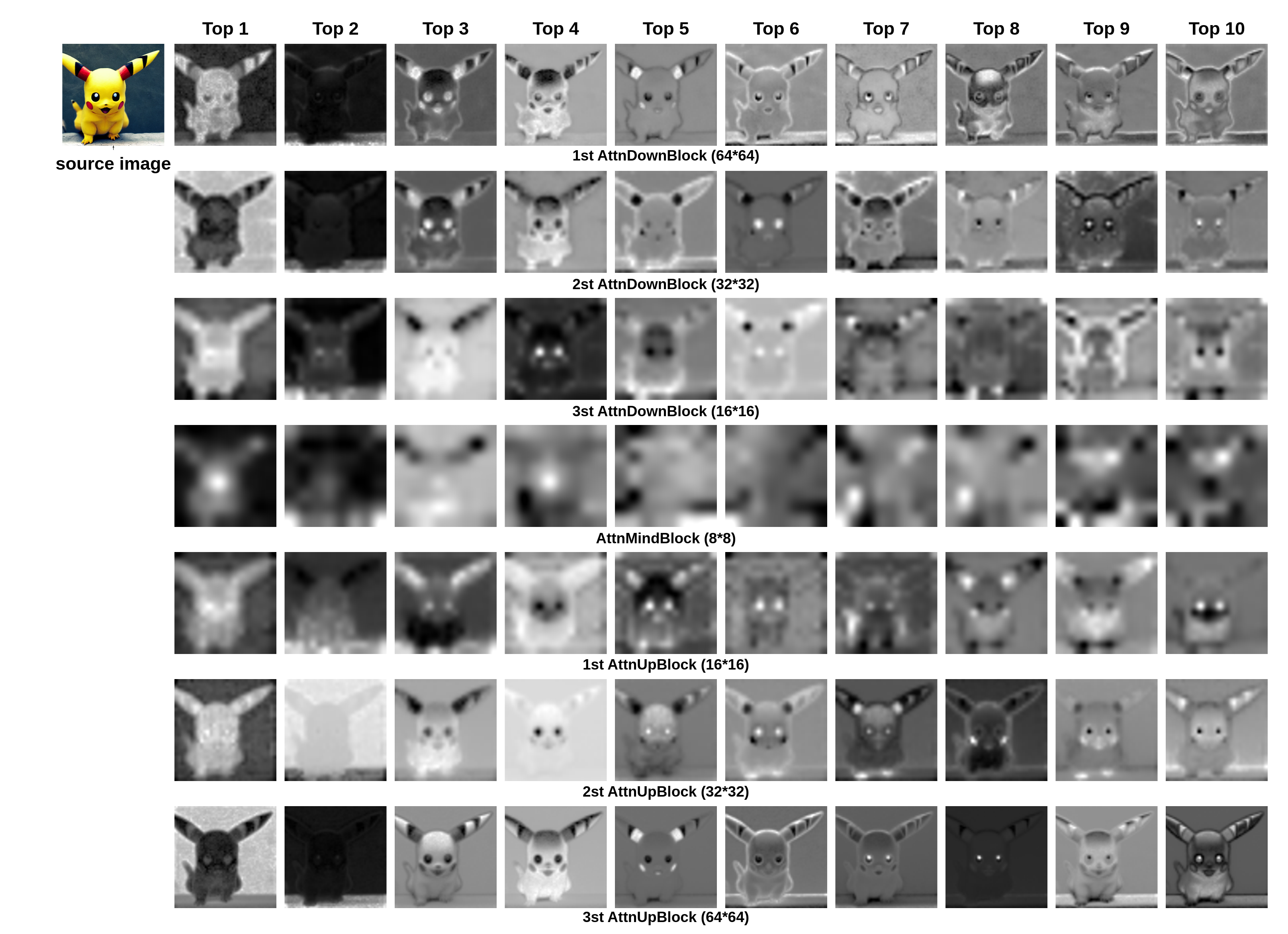}

    \caption{Heatmaps of Self-Attention Maps from Different Attention Blocks in the U-Net of a Stable Diffusion Model. This figure presents heatmaps where each heatmap corresponds to one of the top-10 principal components of a self-attention map, obtained through Singular Value Decomposition (SVD) \cite{wall2003singular}. These components are derived from self-attention maps within various attention blocks of the U-Net architecture. The image analyzed was generated with the prompt: a photo of a Pikachu. For consistent presentation and comparison, all displayed principal component heatmaps have been resized to a uniform resolution of 512×512 pixels. As can be seen, heatmap from higher resolution self-attention maps provide more detail and finer structural information about the depicted objects.}
    \label{fig:self_cross_attention_map}
\end{figure}

\begin{figure}
    \centering
    \includegraphics[width=0.5\textwidth]{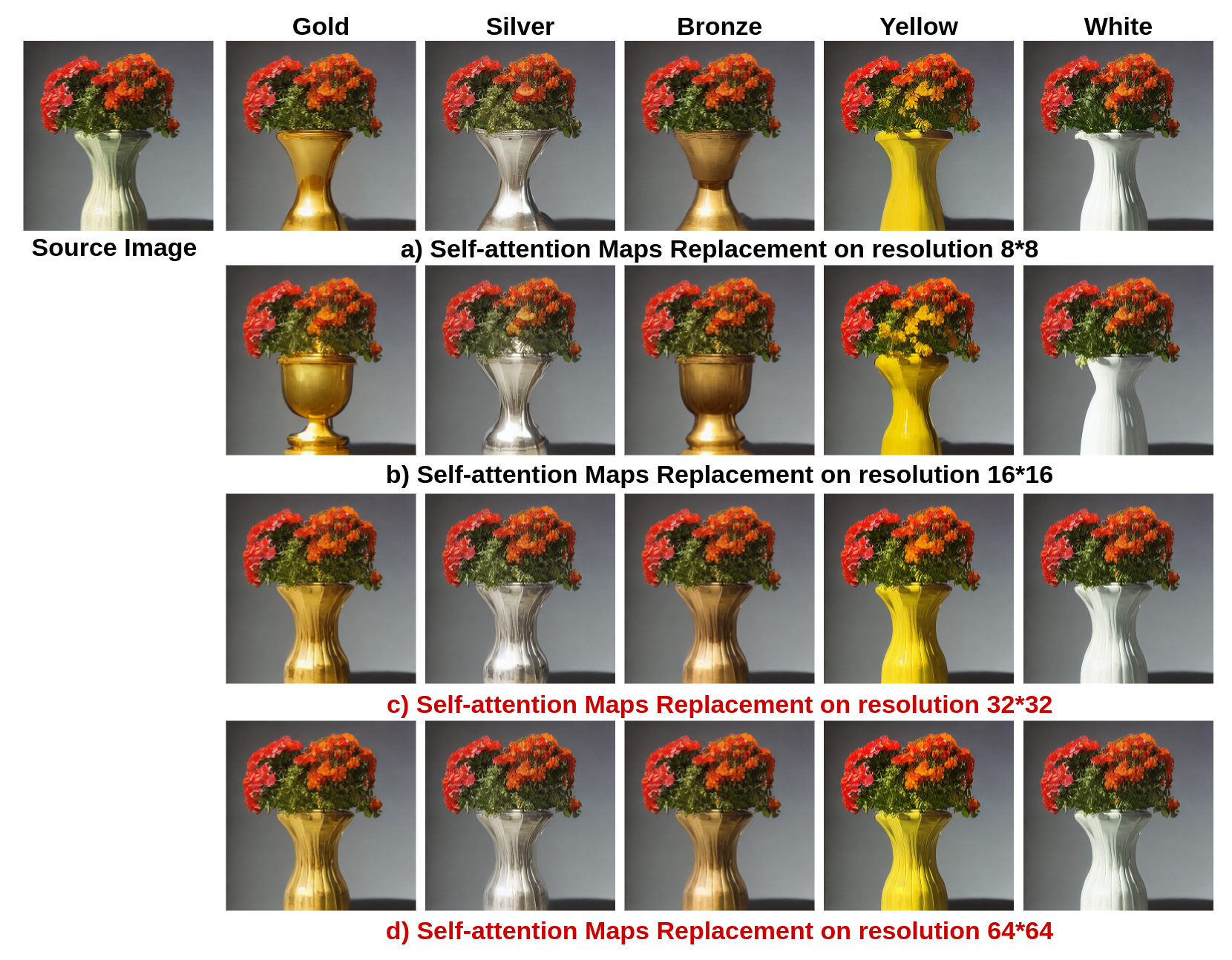}

    \caption{Impact of Varying Self-Attention Map Resolutions on Structural Integrity Preservation. This figure reveals the crucial role of self-attention map replacement resolution in maintaining structural integrity during the denoising process. As illustrated, robust structural preservation can be effectively achieved by replacing self-attention maps of high resolutions (higher than 32*32) throughout the entire denoising process.} 
    \label{fig:p2p_selfattention_replacement_ablation}
\end{figure}

\begin{figure}
    \centering
    \includegraphics[width=0.5\textwidth]{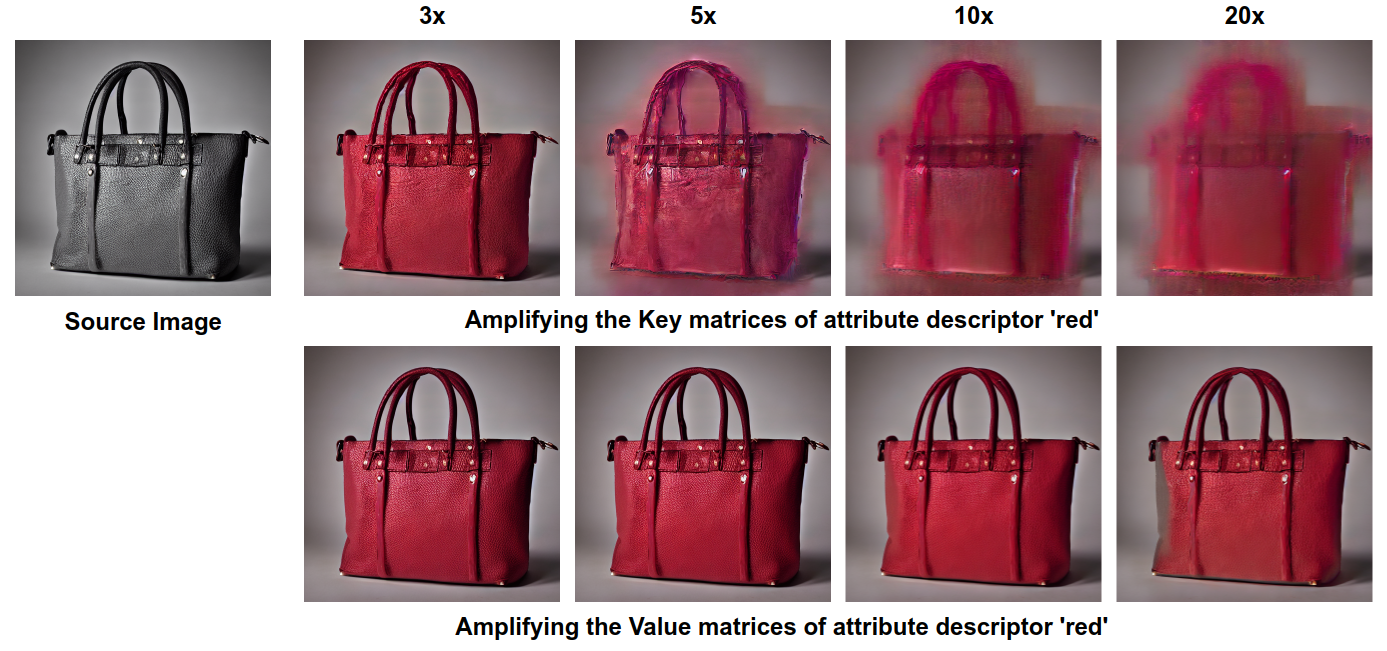}

    \caption{Cross-Attention Key and Value Amplification for Attribute Modifications with Structure Preservation.This figure demonstrates the impact of amplifying an attribute descriptor within the cross-attention layer for attribute modifications, with self-attention map replacement. As shown, amplifying the Value component consistently yields more stable attribute modifications compared to modifying the Key.} 
    \label{fig:cross_attention_key_value_amplify}
\end{figure}
\begin{figure}
    \centering
    \includegraphics[width=0.5\textwidth]{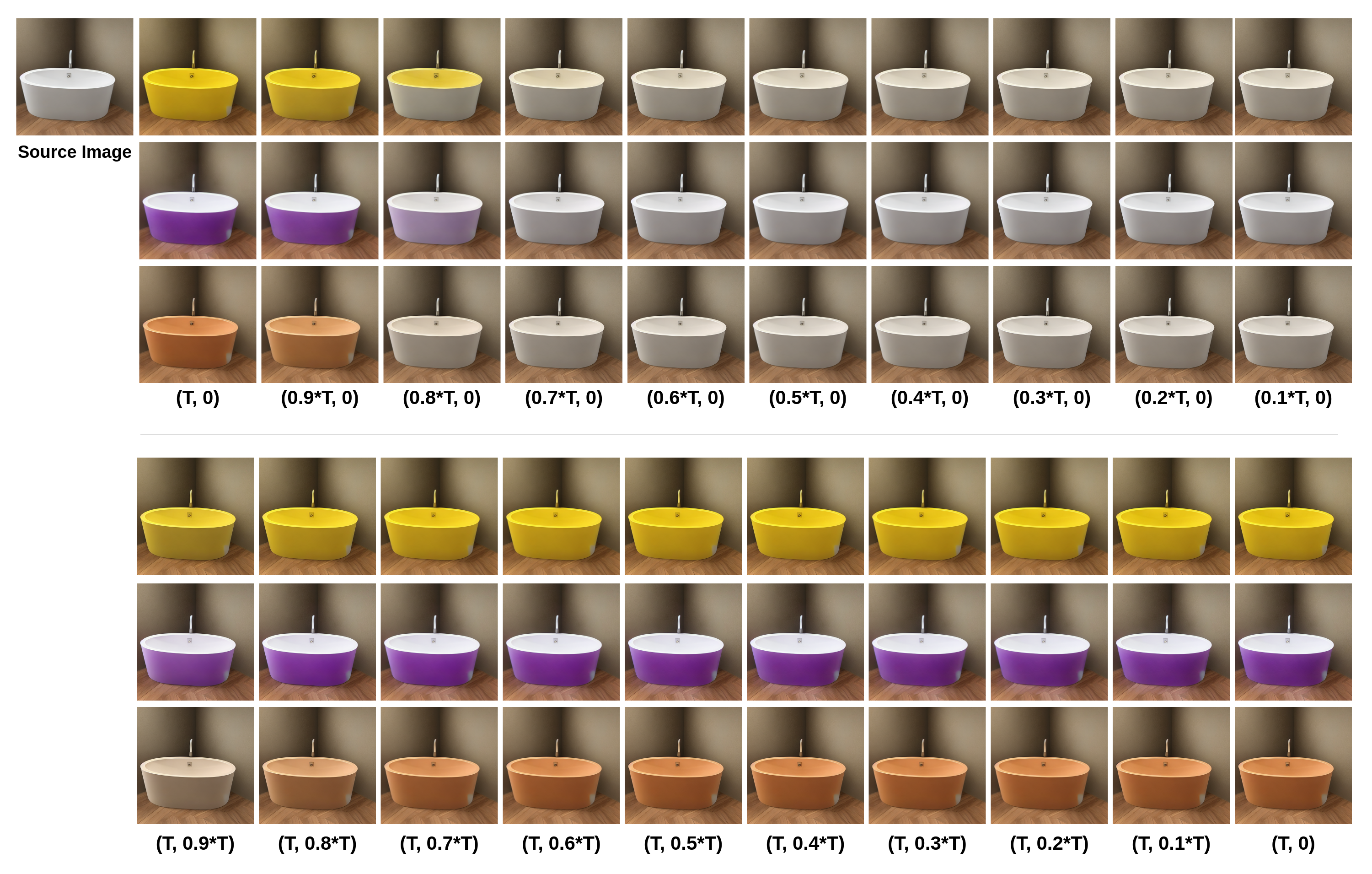}
    \caption{Impact of Cross-Attention Value Amplification Stage on Attribute Modification.This figure presents the results of applying cross-attention value amplification with a constant ratio at different stages of the denoising process. As demonstrated, successful modification of object attributes necessitates the application of attribute amplification from the early stages of the denoising process. Once the attribute of an object change, amplifying the cross-attention value of attribute descriptor in  later stages have minimal impact.} 
    \label{fig:cross_attention_value_amplify_in_denoising_process}
\end{figure}

\begin{figure}
    \centering
    \includegraphics[width=0.5\textwidth]{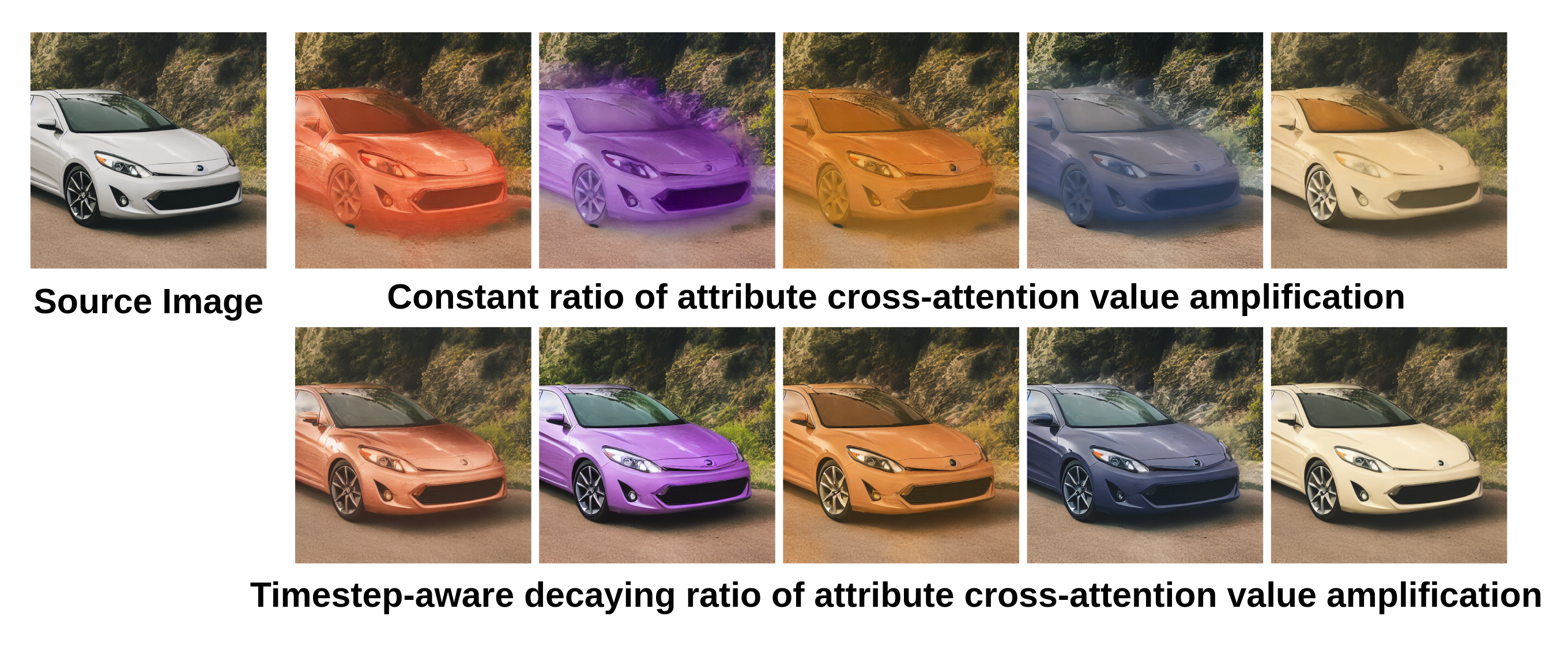}

    \caption{Comparison of Timestep-Aware Decaying versus Constant Amplification Ratios for Cross-Attention Values. This figure presents a comparative analysis of attribute modification results when amplifying the cross-attention value using either a constant ratio or a timestep-aware decaying ratio. As demonstrated, the timestep-aware decaying ratio yields superior results for attribute amplification compared to applying a constant ratio.} 
    \label{fig:constant_ration_amplify_and_timeste_aware_decaying}
\end{figure}

\begin{figure}
    \centering
    \includegraphics[width=0.5\textwidth]{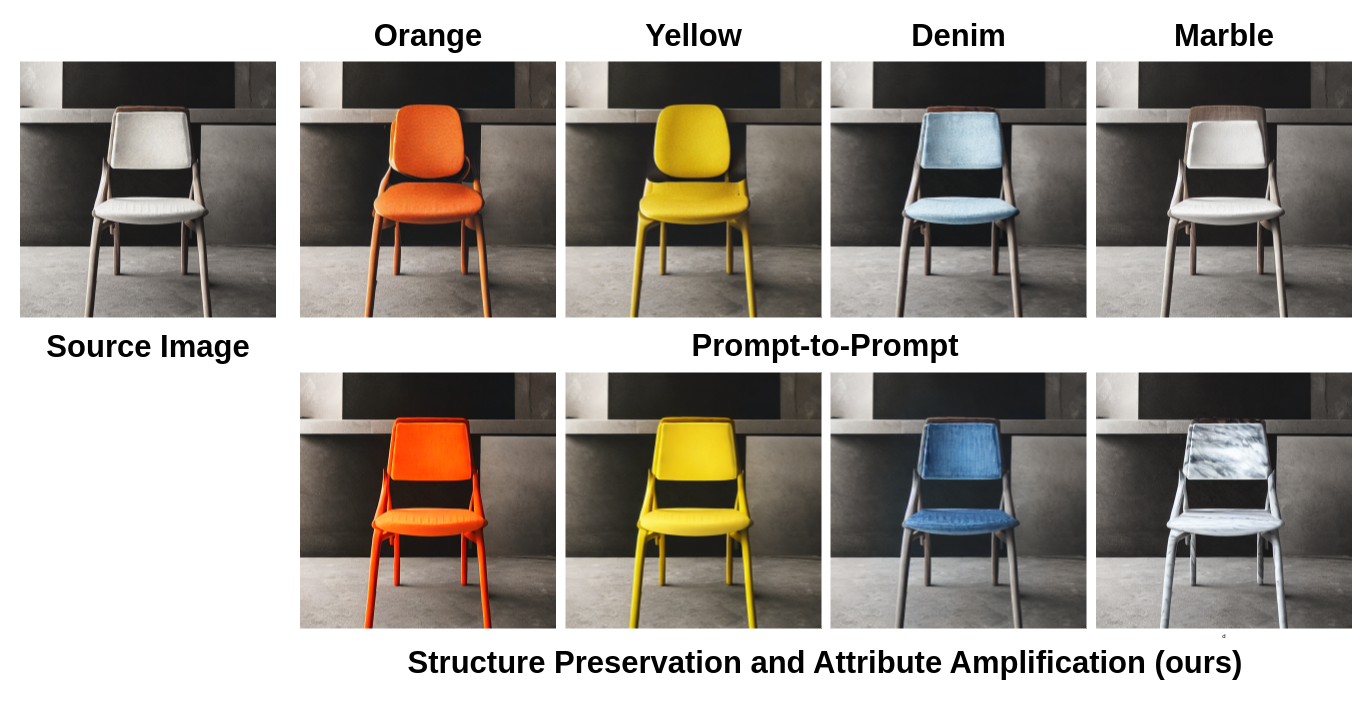}

    \caption{Comparison of Object-Attribute Generation between P2P and the Proposed SPAA Approach. This figure compares the object-attribute generation capabilities of the P2P method against our proposed SPAA approach. As evidenced, SPAA excels at generating objects with varied colors and materials, crucially maintaining their structural integrity.} 
    \label{fig:p2p_selfattention_replacement_crossattention_value_amplify}
\end{figure}

\begin{figure}
    \centering
    \includegraphics[width=0.5\textwidth]{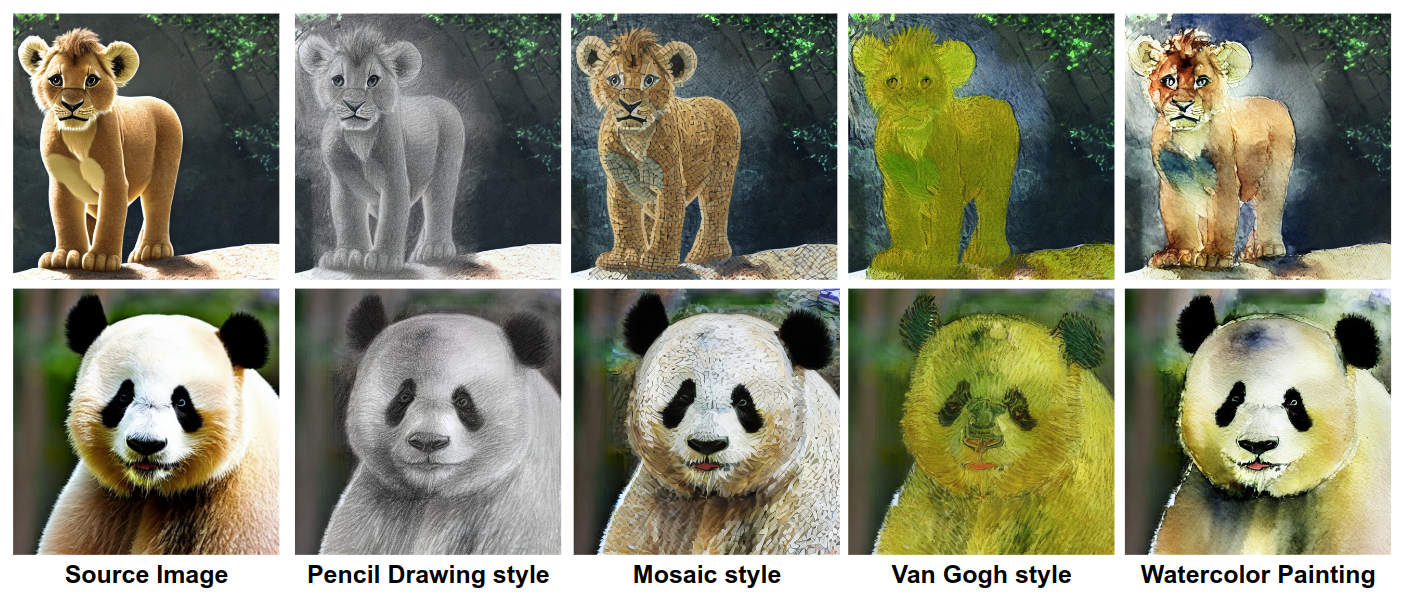}

    \caption{SPAA for object Painting Style Generation.This figure demonstrates the capability of our proposed SPAA method for controlling object painting style. As evidenced, SPAA successfully enables the generation of objects with diverse painting styles while maintaining their original structural integrity.} 
    \label{fig:style_transfer}
\end{figure}

\begin{figure}
    \centering
    \includegraphics[width=0.5\textwidth]{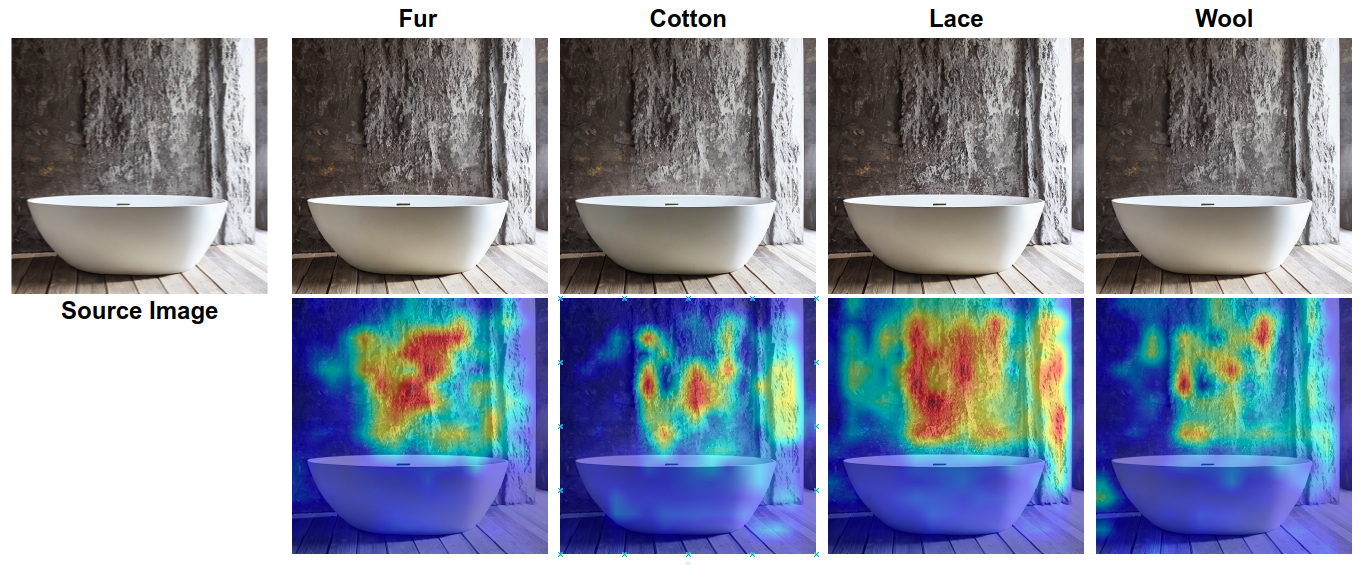}
    \caption{Illustrative Failure Cases of Attribute Modification with Our SPAA Approach. This figure presents examples where attribute modification, attempted using our Structure Preservation and Attribute Amplification (SPAA) method, was unsuccessful. These failures primarily occur when the cross-attention map of the attribute descriptor is not adequately distributed across the target subject, leading to incomplete or incorrect attribute application.} 
    \label{fig:failure_case_with_spaa}
\end{figure}

\begin{figure}
    \centering
    \includegraphics[width=0.5\textwidth]{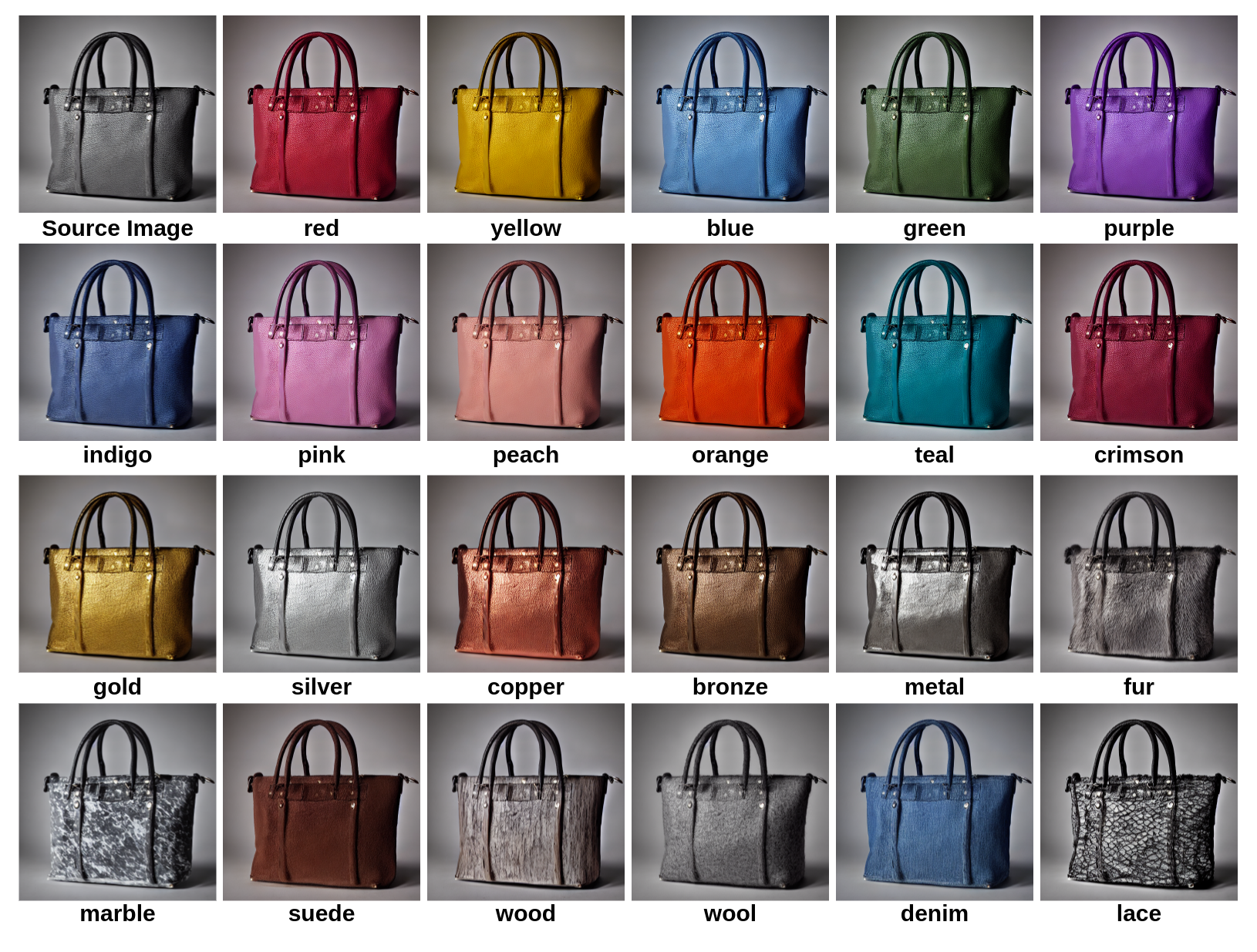}
    \caption{Examples of Source and Target Image Pairs for Object Attribute Generation. This figure showcases various object attribute variations of a handbag. The color variations including red, yellow, blue, green, purple, indigo, pink, peach, orange, teal, crimson, gold, silver, copper and bronze hues. The material transformations including metal, fur, marble, suede, wood, wool, denim and lace. Observe the consistent structure preservation across all generated variations, along with the accurate of the specified attributes.} 
    \label{fig:source_target_sample}
\end{figure}

\begin{figure}
    \centering
    \includegraphics[width=0.5\textwidth]{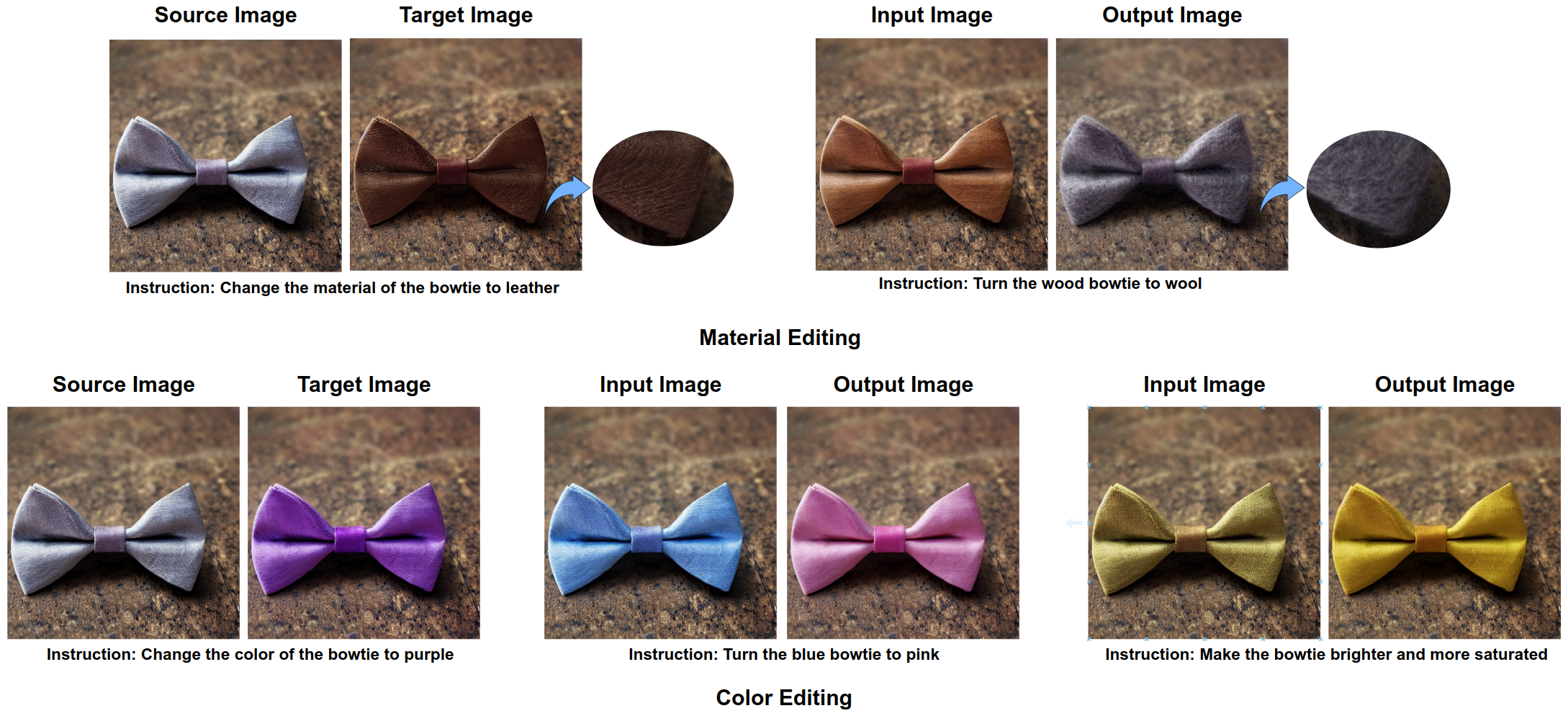}

    \caption{Examples of Instructional Triple Data Samples.This figure presents various examples of instructional triple data samples, demonstrating the diversity of instructions encompassed within our dataset. These distinct instruction types are specifically designed to enhance both the diversity and efficacy of the InstructAttribute model.} 
    \label{fig:instrcution_type}
\end{figure}

Our aim is to create a generalized model for fine-grained object attributes editing. However, due to the lack of high-quality fine-grained attribute variation datasets for training and evaluation, the advancement of diffusion models in this field has been notably limited. Manually generating datasets with tools like Photoshop is prohibitively time-consuming, and publicly available synthetic datasets, such as HQ-Edit \cite{hui2024hq} and UltraEdit \cite{zhao2024ultraeditinstructionbasedfinegrainedimage}, frequently focus on the most common colors (white, black, red, yellow, blue, green, and purple, etc.) and materials (like wood, metal, iron, etc.).  To generate high-quality fine-grained object attribute variation image pairs, we proposed the Structure Preservation and Attribute Amplification method, which is capable of generating variations in color, material, and style for the same object while preserving its structure. Then we introduce a data collection pipeline that utilizes advanced foundation models LLaVa \cite{liu2023llava}, DINO \cite{caron2021emerging} , Grounding Dino \cite{liu2023grounding}, and SAM \cite{kirillov2023segment} to minimize the reliance on human feedback for the selection of the target attribute image. Finally, utilizing the in-context learning capabilities of GPT-4o \cite{hurst2024gpt}, we systematically generate a diverse set of instruction templates for object attribute modification.  The pipeline of our dataset collection engine is given in Fig.~\ref{fig:pipeline}. 

\textbf{Structure Preservation and Attribute Amplification.} Generating plausible fine-grained attribute variations for a subject poses a significant challenge, as it requires precise control to modify object attributes while maintaining the structural integrity of the subject and the contextual consistency of the background. One approach for generating fine-grained attribute variations involves fine-tuning the Stable Diffusion model to associate a specialized token with a target attribute concept, thereby enabling precise control over object characteristics. For example, ColorPeel \cite{butt2024colorpeel} enhances color-specific control in diffusion models by disentangling color and shape characteristics during image generation through the learning of color concepts. Material Palette \cite{lopes2024material} fine-tune the diffusion model to learn the material depicted as a token from a single input image. This material concept is used to generate a texture image and then extract the Physics-Based Rendering (PBR) materials. While concept learning in Stable Diffusion enables remarkable customization for generating or manipulating specific visual concepts, it faces significant practical challenges, primarily due to substantial computational overhead (e.g., GPU memory demands) and excessively time-consuming workflows. In contrast, training-free methods, such as Prompt-to-Prompt (P2P), leverage textual descriptors to directly define object attributes without requiring model retraining. However, P2P may inadvertently alter the structure of the object or struggle to to modify the attributes of the object (see an example in the first row of Fig.~\ref{fig:p2p_selfattention_replacement_crossattention_value_amplify}).  For structural preservation, MasaCtrl \cite{cao2023masactrl} and Free \cite{liu2024towards} have shown that self-attention maps derived from textual prompts encode structural and layout information from generated images. As can be seen in Fig.~\ref{fig:self_cross_attention_map}, which
visualizes the average self-attention maps of the Pikachu from different attention blocks over all time steps. Analysis of self-attention maps reveals that distinct U-Net blocks encode structural information at varying resolutions. Lower-resolution blocks (e.g., below 16×16 px) primarily capture object shape and contour, whereas higher-resolution blocks (e.g., 32×32, 64×64 px) encode finer structural details and textural information. Given this insight, we contend that the inability of P2P \cite{hertz2022prompt} to preserve object structure stems from its exclusive reliance on manipulating self-attention maps in low-resolution layers and an ablation study is given in Fig.~\ref{fig:p2p_selfattention_replacement_ablation}.  As can be seen, the structural preservation problem can be solved by replacing the target self-attention maps $M^{T}_{self}$ by source self-attention maps $M^{S}_{self}$ on high resolutions throughout the denoising process. For the failure to modify object attributes, we argue that it stems from insufficient guidance provided by text-based attribute descriptors during the denoising process. Specifically, these descriptors lack the semantic granularity necessary to steer attribute-specific features in the cross-attention layers. A straightforward strategy to address this limitation involves amplifying the cross-attention Key or Value matrices associated with the target attribute descriptor in the prompt. However, as demonstrated in Fig.~\ref{fig:cross_attention_key_value_amplify}, scaling the key matrices, which directly govern the cross-attention map, often induces unintended structural distortions in the generated subject. In contrast, adjusting the value matrices, which encode attribute-specific feature representations, achieves more stable and coherent transformations while preserving the geometric integrity of the object. Therefore, we propose a cross-attention Value amplification method to strengthen the effect of attribute modification. To investigate the impact of cross-attention value amplification, we conducted an ablation study across ten stages of the denoising process, amplifying attribute descriptor cross-attention values on different sequence of stages, (see Fig.~\ref{fig:cross_attention_value_amplify_in_denoising_process}). To effectively modify an object's attributes, attribute amplification must be applied from the very beginning of the denoising process. Once an object's attribute has been altered, amplifying the cross-attention value of its descriptor in later stages has minimal impact. In practice, we observed that a time-step-aware decaying ratio for amplifying attribute cross-attention values yielded better results than a constant ratio. We argue that a high amplification ratio in the early stages can introduce excessive information, particularly for certain objects, see an example in Fig.~\ref{fig:constant_ration_amplify_and_timeste_aware_decaying}.  Together, we refer to this combined method as \textbf{Structure Preservation and Attribute Amplification}.  So, our denoising process is:
\[
z^{T}_{t\text{-}1}= \text{DM}(z^{T}_t, P^T, t, \lambda, R)\{M^{T}_{cross} * V^{T^*}_{cross}\} \{M^{T}_{self} \gets M^{S}_{self}\}
\]
where $\lambda$ is the decaying ratio, $R$ is the initial amplification ratio,  $V^{T^*}_{cross}$ represents the updated cross-attention value of the target prompt $P^T$ in the denoising process, which is given by:
\[
V^{T^*}_{cross} = 
\begin{cases}
(1-\lambda)* t * R *V^T_{cross_i}, \space\text{$if P^T_i \space\in \space attrs$} \\
 V^T_{cross_i}, \space\text{$other$}
\end{cases} 
\]
$P^T_i$ represent the $i$ word in the target prompt $P^T$ and $attrs$ is a list of attribute discriptors in $P^T$. The pseudocode of the attribute variation generation algorithm is given in Algorithm 1. We also observed that our SPAA method is effective for the transformation of the object painting style, as demonstrated in Fig.~\ref{fig:style_transfer}. However, painting styles (e.g., watercolor, van Gogh, mosaic) are inherently holistic, shaping the overall aesthetic of an image, so the object-specific style modifications are excluded from this study.

\begin{algorithm}[!h]
\caption{structure preservation and Attribute Amplification for object-attribute variation image generation}
\label{alg: ALG1}

    \begin{flushleft}
        \hspace{0cm}  \textbf{Input:} A source prompt $P^S$,  A target prompt $P^T$, an attribute discriptor list $attrs$ in $P^T$    \\
        \hspace{0cm}  \textbf{Output:}  A source image $I^{S}$, target image $I^{T}$  \\
    \end{flushleft}
    
    \begin{algorithmic}[1]
        \State $z^{s}_T \in N (0, 1)$, a unit Gaussian random value sampled with random seed
        \State  $R$, a initial ratio to amplify the Value matrix of attribute discriptor
        \State  $\lambda$, a decaying ratio
        \State  $z^T_T \gets z^T_S$
            
    \For {$t = T, T-1, ..., 1$}   
            \State $M^{S}_{cross_{t}}, M^{S}_{self_{t}}  =\text{DM}(z^{S}_t, P^S, t)$
            \State $V^T_{cross_t},M^{T}_{cross_{t}}, M^{S}_{self_{t}} =\text{DM}(z^{T}_t, P^T, t)$
            
            \For{$i = 0, 1, ..., len(P^T)$}
                        \If {$ P^T_i\in attrs$}
                            \State $V^{T^*}_{cross_i} = (1-\lambda)* t * R *V^T_{cross_i}  $
                        \Else
                            \State $V^{T^*}_{cross_i} = V^T_{cross_i}$                               
                        \EndIf
                    \State 
            \EndFor
            
$z^{T}_{t\text{-}1}= \text{DM}(z^{T}_t, P^T, t)\{M^{T}_{cross} * V^{T^*}_{cross}\} \{M^{T}_{self} \gets M^{S}_{self}\}$
    \EndFor            
    \State $ I^{S} \gets LDMImageDecoder(z^{S}_{1})$ 
    \State $ I^{T} \gets LDMImageDecoder(z^{T}_{1})$ 
    \State \Return $I^{S}$, $I^T $
    \end{algorithmic}    
\end{algorithm}

\begin{table}
 \caption{Attribute modification success rate across different components. Our method achieves the highest success rate under the same target attribute selection criterion.}
 \label{tab:t1}
    \centering
    \begin{tabular}{|c|c|c|}
    \hline
    Method& $LLaVa\uparrow$&$LLaVa + DS\uparrow$\\
    \hline
    P2P&  47.5\%& 7.3\%\\
    \hline
 + Structure Preservation& 36.9\%& 22.4\%\\
    \hline
 + Attribute Amplification& \textbf{74.3\%}& 5.2\%\\
     \hline
    Ours& 72.5\%& \textbf{61.3\%}\\
     \hline
\end{tabular}

\end{table}

\textbf{Target attribute image selection.} While the Structure Preservation and Attribute Amplification method significantly increases the success rate in generating attribute variation images, we notice that attribute modification can occasionally fail due to misdistribution of the cross-attention map of the attribute descriptor and the object (see an example in Fig.~\ref{fig:failure_case_with_spaa}). To select the desired image for the modification of the attribute, we employ LLaVa \cite{liu2023llava} to verify whether the attribute of the subject in the target image accurately corresponds to the designated attribute.  This is achieved by posing the query: "What \{\} does the \{\} appear to be? \{\}? Answer yes or no." We retain target images that receive a "yes" response. During the attribute verification process, the first placeholder is replaced with the attribute type, such as color or material, the second placeholder is replaced with the subject name, and the third placeholder is replaced with the target attribute descriptor. For those retained target images, the structure and texture of the object in the attribute modification scenario should remain largely unchanged.  To ensure this, we calculate the DINO grayscale score \cite{caron2021emerging} between the source and target images to assess semantic similarity, discarding targets with a score below a certain threshold (0.90 for material and 0.98 for color attribute). Under this selection criterion, we conducted an ablation study to assess the effectiveness of our structure preservation and attribute amplification method, demonstrating the contribution of each component. As can be seen in Table \ref{tab:t1}, with the same selection criterion, our structure preservation and attribute amplification method yield the highest success rate. 

\textbf{Attribute Leakage filtering.} Given a source image and its corresponding target attribute image, we compute the absolute pixel-wise difference between them. The DINO grounding \cite{liu2023grounding} is used to detect the object in the source image, and the SAM \cite{kirillov2023segment} is used to generate a segmentation mask for the object. This object mask is then inverted to obtain the background mask. By applying the background mask to the absolute difference image, we isolate the changes occurring in the background region. We then count the number of pixels with non-zero values in this region. If the count exceeds a predefined threshold, the target attribute image is considered to exhibit attribute leakage into the background and is therefore discarded. In this study, the threshold is set to 50. An example of a source image and the target attribute images is given in Fig.~\ref{fig:source_target_sample}.

\textbf{Object-Attribute Editing instruction template.}  To construct the ATTRIBUTE dataset, initially, we collected class names from established datasets COCO \cite{lin2014microsoft}, ImageNet \cite{deng2009imagenet}, Open Images Dataset \cite{kuznetsova2020open} and VisualGenome \cite{krishna2017visual}, resulting in a comprehensive list of more than 3,300 subjects, including foods, animals, plants, toys, clothing, transportations, daily necessities, natural objects , and more. Subsequently, for each subject, we generate the source images using 19 distinct prompt templates combined with randomly selected seeds. As altering a subject's material inherently modifies its structure and texture, whereas color modification should minimally affect these properties, we divide the attribute dataset into a color-editing dataset and a material-editing dataset. To automatically generate the instructional text prompt for each pair of source-target images, we use the ability of GPT-4o \cite{hurst2024gpt} to generate a range of prompt template. Firstly, we randomly select 200 pairs of source and target images, where the object in the target images has varying attributes. Using the in-context learning ability of GPT-4o \cite{hurst2024gpt}, we ask the model to identify the differences between each pair and to explain how we can convert one image to another image. Based on its responses, we develop two types of editing instruction templates to enhance attribute editing diversity: (a) instructions for transforming the source image into the target image, and (b) instructions for generating target images with different attributes. The first type of instruction guides the model in altering intrinsic object attributes to entirely different ones, while the second enables transitions between different attributes of the same object. For color modification, we add a new category of editing instruction templates focused on generating target images within the same color hue. This instruction enhances the model's ability to fine-tune colors within the same hue by adjusting brightness and saturation levels.  For each pair of source-target images, we randomly select one instruction template to create the triple training dataset. An example of a dataset sample is shown in Fig.~\ref{fig:instrcution_type}.

%% file: sec/4_InstructionAttribute.tex
\section{ InstructionAttribute}
\label{sec:InstructionAttribute}

\textbf{Instructional Supervised Training.} With the object attribute dataset, we adopt the instructional fine-tuning approach of InstructPix2Pix \cite{brooks2023instructpix2pix} to train our method to edit the attributes of the object. This approach utilizes a pre-trained Latent Diffusion Model (LDM) \cite{rombach2022high} as its primary framework, implementing a two-stage synthesis process that performs diffusion and denoising in a compressed latent space while ensuring high-quality results. During training, a pre-trained autoencoder with an encoder $E$  and a decoder $D$ is used to first map the edited image $x$ to its latent representation $z = E(x)$. The model then learns a network $\theta$ to predict the noise added to the noisy latent $z_t$, conditioned on both the image $c_I$ and the text instruction $c_T$, where $c_T$ is concatenated with $z_t$ and $\epsilon(c_I)$. The network $\theta$ , which is initialized with Stable Diffusion weights, is fine-tuned to minimize the latent diffusion objective. \[
 E_{\epsilon(x), \epsilon(c_I), c_T, \epsilon \sim \mathcal{N}(0, I), t}[||\epsilon - \epsilon_\theta(z_t, t, \epsilon(c_I), c_T)||^2_2]
 \]
 
\textbf{Training Dataset. } For color editing, we focus on 43 colors that are frequently observed in daily life, which including amethyst, azure, beige, black, blue, bronze, brown, camel, copper, coral, cream, crimson, cyan, emerald, gold, gray, green, indigo, khaki, lime, magenta, maroon, navy, olive, orange, peach, pink, plum, purple, red, rose, salmon, silver, slate, tan, taupe, teal, tomato, turquoise, violet, white, wine, yellow. In practice, we find that initially the cross-attention value amplifying ratio of the color descriptor to 5 and gradually reducing it by 0.1 per denoising step until it reaches 1.0 yields the optimal target color image success rate. Finally, we generate more than 0.82M pairs of source-target color image pairs after the image selection process. Through the instructional training dataset generation process, the color editing scenario, a triple dataset consisting of 1.8M input-output image pairs is generated, along with the corresponding edit instructions.  For material editing, we examine 14 commonly observed materials in daily life, including cotton, glass, marble, plastic, velvet, denim, lace, mesh, wood, fur, leather, metal, suede, and wool. For the target material subject generated, we find that initializing the cross-attention Value amplifying ratio at 10 of the material descriptor and progressively reducing it by 0.2 per denoising step until reaching 1.0 achieves the highest success rate for generating target material images. In total, we generated over 0.31M source-target material image pairs after the image selection process. Using those image pairs, we create a triple dataset for the material editing scenario, consisting of 0.7M input-output image pairs and their corresponding editing instructions. 

%% file: sec/5_experiment.tex
\section{Experiment}
\label{sec:experiment}

\begin{figure*}[!ht]
    \centering
    \includegraphics[width=1.0\textwidth]{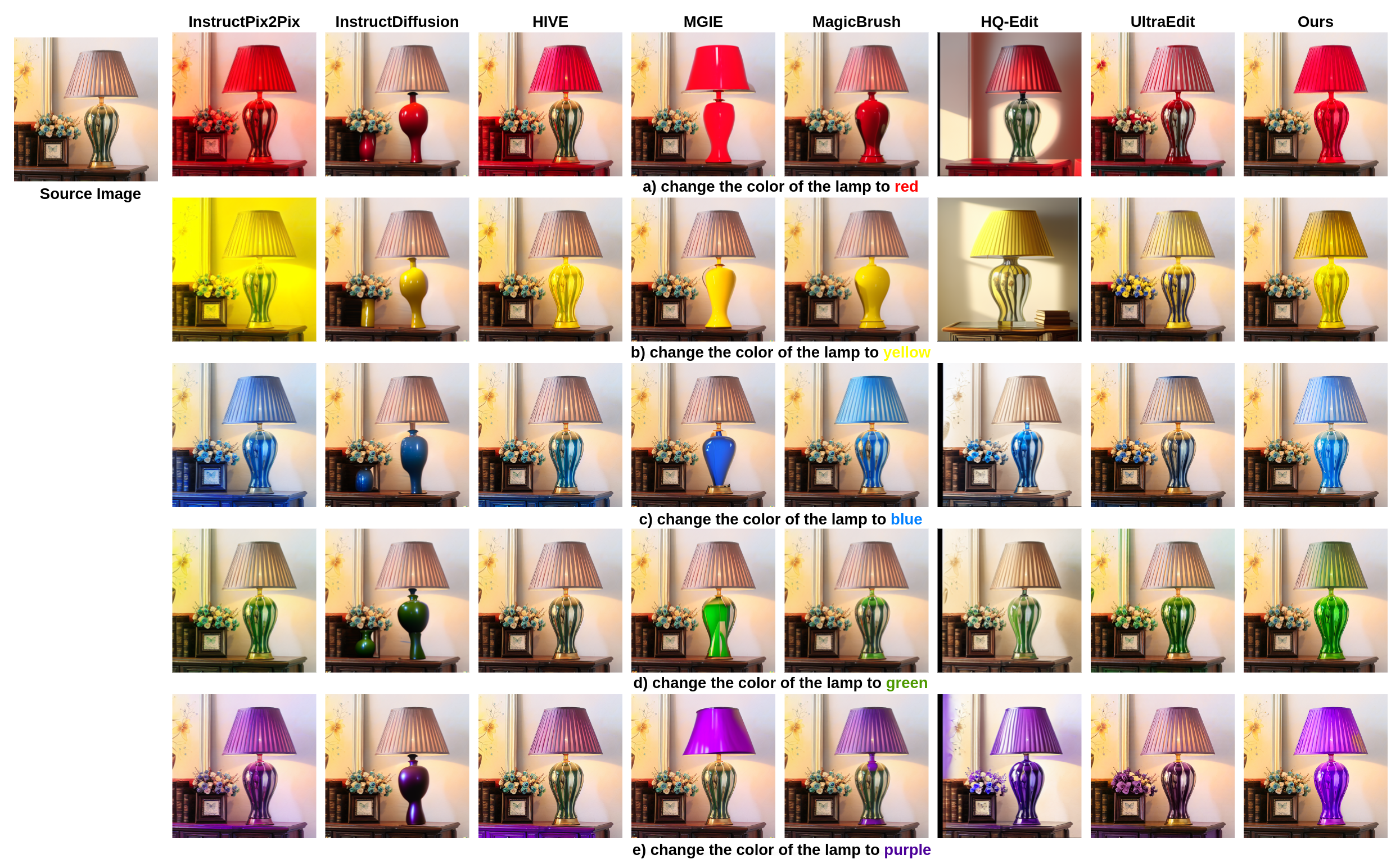}

    \caption{Comparative Results of Instruction-Based Color Alteration on Common Colors. This figure presents a comparative analysis of color alteration capabilities using instruction-based editing methods. For the color modification task, InstructDiffusion (InsDiff) \cite{geng2024instructdiffusion}, MGIE \cite{fu2023guiding}, and MagicBrush \cite{Zhang2023MagicBrush} distorted the structure of the lamp. InstructPix2Pix (IP2P) \cite{brooks2023instructpix2pix} and HIVE \cite{han2024ace} exhibited attribute leakage into the background area. HQ-Edit \cite{hui2024hq} altered the background and even removed the photo album. UltraEdit \cite{zhao2024ultraeditinstructionbasedfinegrainedimage} failed to change the lamp's color when attempting to modify it to red or blue. Our method effectively modifies the subject's color while preserving both its structural integrity and background information.}
    \label{fig:qualitative_comparision_with_other_method_common_seem_color}
\end{figure*}

\begin{figure}[!ht]
    \centering
    \includegraphics[width=0.5\textwidth]{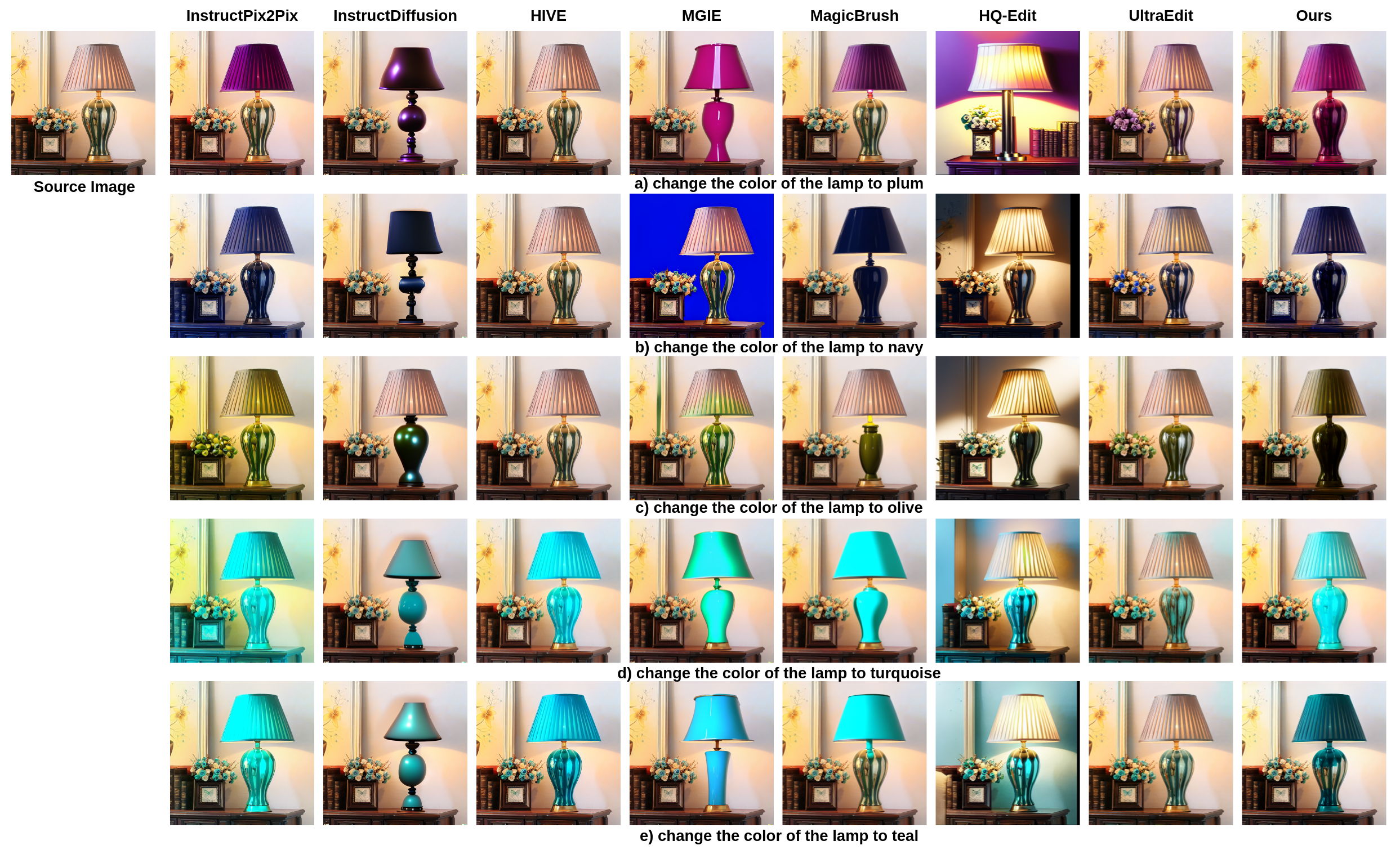}

    \caption{Comparative Results of Instruction-Based Color Alteration in Less Common Hues. This figure presents a comparative analysis of color alteration capabilities using instruction-based editing methods on a selection of less commonly encountered colors. For the less common colors seen, InstructDiffusion (InsDiff) \cite{geng2024instructdiffusion}, MGIE \cite{fu2023guiding}, and MagicBrush \cite{Zhang2023MagicBrush} distorted the structure of the lamp. InstructPix2Pix (IP2P) \cite{brooks2023instructpix2pix} exhibited attribute leakage into the background area when changing the color to navy and turquoise. HQ-Edit \cite{hui2024hq} altered the background. HIVE \cite{han2024ace}  and UltraEdit \cite{zhao2024ultraeditinstructionbasedfinegrainedimage} failed to change the lamp color when attempting to modify it to plum, navy, olive, turquois and teal. }
    \label{fig:qualitative_comparision_with_other_method_less_common_color}
\end{figure}

\begin{figure*}[!ht]
    \centering
    \includegraphics[width=1.0\textwidth]{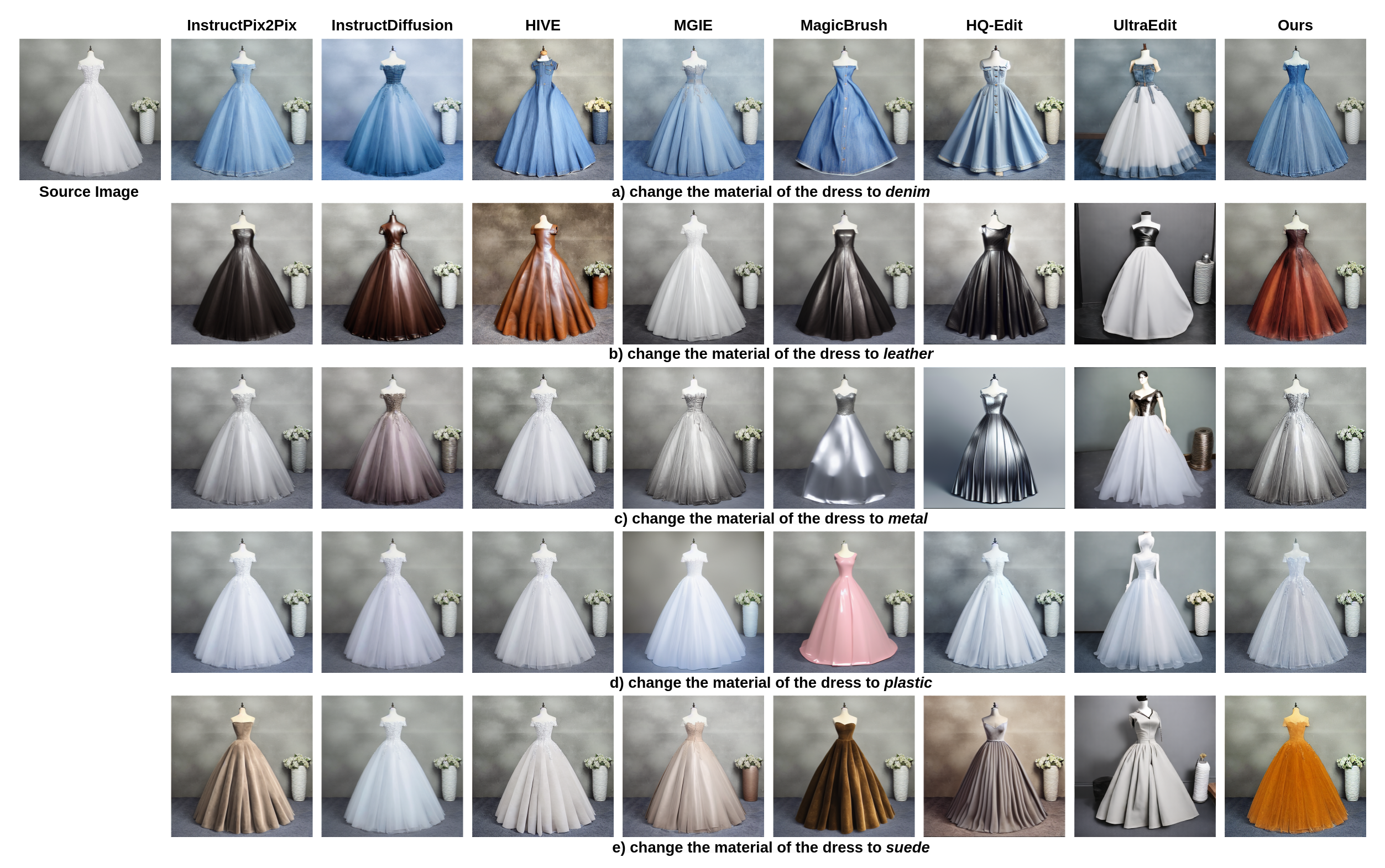}

    \caption{Comparative Material Alteration with Instruction-Based Methods. This figure presents a comparative analysis of material alteration capabilities using various instruction-based editing methods. As can be seen, InstructPix2Pix (IP2P) \cite{brooks2023instructpix2pix}, Instructdiffusion (InsDiff) \cite{geng2024instructdiffusion}, HIVE \cite{han2024ace}, MGIE \cite{fu2023guiding}, MagicBrush \cite{Zhang2023MagicBrush}, HQ-Edit \cite{hui2024hq} and UltraEdit \cite{zhao2024ultraeditinstructionbasedfinegrainedimage} distorted the structure of the dress. Instructdiffusion (InsDiff) \cite{geng2024instructdiffusion}, HIVE \cite{han2024ace}, UltraEdit \cite{zhao2024ultraeditinstructionbasedfinegrainedimage} exhibited attribute leakage into the background. Our method achieves the optimal balance between modifying the material and preserving the subject's original structural. } 
    \label{fig:qualitative_comparision_with_other_method_material}
\end{figure*}

 \begin{figure*}[!ht]
    \centering
    \includegraphics[width=1.0\textwidth]{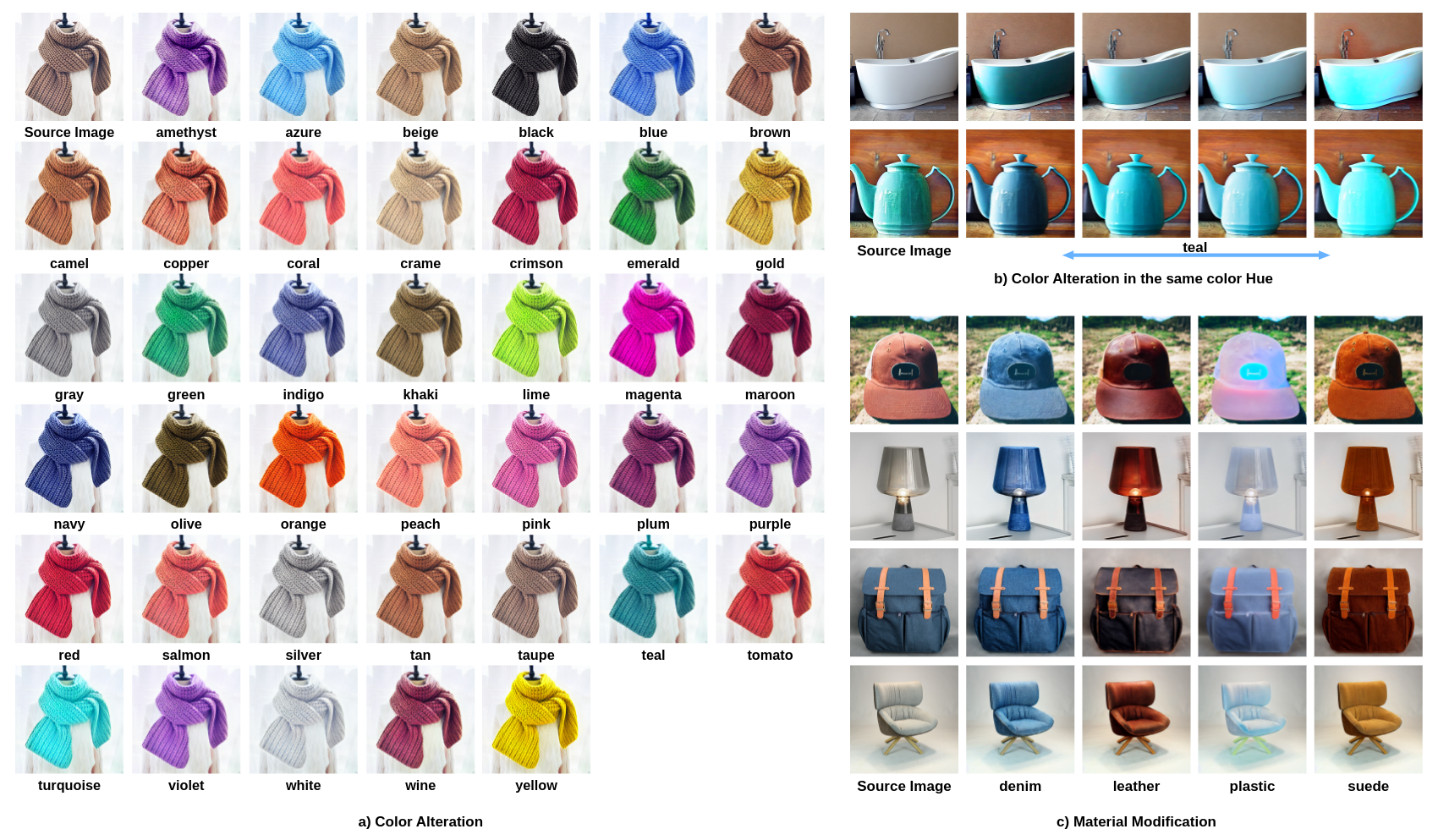}

    \caption{Results of attribute alterations using our InstructAttribute. a) This figure demonstrates the ability of our method in performing precise color adjustments.  b) This figure demonstrates our method's capability to modify an object's color while strictly maintaining its hue. c)  This figure illustrates the results of material modification using our method.} 
    \label{fig:attribute_change}
\end{figure*}

\textbf{Implementation details.} We follow the instructional fine-tuning method in InstructPix2Pix and initialize the model with StableDiffusion \cite{rombach2022high} checkpoints V1.5. During training, we set the image resolution to 256, the total training steps to 20000, and the learning rate to 1.0e-4. During the editing process, we set the image resolution to 512, the image guidance scale to 1.5, the instruct guidance scale to 7.5, and the number of inference steps to 100.

\textbf{Baseline models.}  We make comparisons with the latest instruction-based image editing methods, including InstructPix2Pix (IP2P) \cite{brooks2023instructpix2pix}, Instructdiffusion (InsDiff) \cite{geng2024instructdiffusion}, HIVE \cite{han2024ace}, MGIE \cite{fu2023guiding}, MagicBrush \cite{Zhang2023MagicBrush}, HQ-Edit \cite{hui2024hq} and UltraEdit \cite{zhao2024ultraeditinstructionbasedfinegrainedimage}. To maintain reproducibility and fairness, we employed the default hyper-parameters provided in the official implementations. 

\textbf{Qualitative evaluation.} For the color modification task, as illustrated in Fig.\ref{fig:qualitative_comparision_with_other_method_common_seem_color}, prevalent colors such as red, yellow, blue, green, and purple, methods such as IP2P and HIVE may unintentionally alter the color of the background elements while adjusting the subject's color (e.g., the color of the table is affected when changing the lamp to red, blue, or purple), while HQ-Edit change the structure of background area,  for example, it remove the books and album when change the color of the lamp to red and yellow. Methods such as InsDiff, MGIE, and MagicBrush may compromise structural coherence during color transformations (e.g., when the lamp's color is transformed to red), leading to unintended geometric distortions. Although UltraEdit demonstrates superior structural preservation, its ability to achieve precise color transitions for target objects remains inconsistent, frequently failing to meet desired chromatic outcomes. In contrast, our InstructAttribute achieves superior precision in color modification while preserving both the structural integrity of objects and the consistency of background elements, ensuring that no unintended alterations occur during the editing process.  Our method achieves consistent performance even for less common colors, such as plum, navy, olive, turquoise, and teal, as demonstrated in Fig.~\ref{fig:qualitative_comparision_with_other_method_less_common_color}.  For material modification tasks, as demonstrated in Fig.~\ref{fig:qualitative_comparision_with_other_method_material}, current instructed-based methodologies, including IP2P, InsDiff, HIVE, MGIE, MagicBrush, HQ-Edit, and UltraEdit, exhibit limitations in the execution of precise material transformations while frequently inducing unintended structural distortions or attribute leakage to the background area. In contrast, our proposed approach achieves a refined equilibrium between the modification of the target material and the rigorous preservation of the geometric and structural fidelity of the object, establishing itself as the benchmark for precision in comparative evaluations. Furthermore, Fig.~\ref{fig:attribute_change} a) shows that our model is capable of accurate and precise color adjustments while maintaining the structural integrity of the subject. Fig.~\ref{fig:attribute_change}  b) reveals that our method is able to adjust an object's color within the same hue by modifying its saturation and brightness, preserving the original hue while achieving targeted chromatic variations. Fig.~\ref{fig:attribute_change} c) illustrates that our method ensures a consistent effect of material transformation on various subjects while preserving the structure of the object and the background information. 

\begin{table*}
  \caption{Quantitative evaluation for the color alteration task and the material modification task with different methods. }
  \label{tab:t2}
    \centering
\begin{tabular}{|c|c|c|c|c|c|c|c|c|}
    \hline
    &      \multicolumn{5}{c}{\underline{\hspace{2.8cm}Color Alteration\hspace{2.8cm}}}&\multicolumn{3}{c}{\underline{\hspace{0.7cm}Material  modification\hspace{0.7cm}}}  \\
    
 Method&      $DS\uparrow$&$SSIM\uparrow$&$CS\uparrow$&$L1^{Hue}_{obj} \downarrow$&$LPIPS_{bg} \downarrow$ &$DS\uparrow$& $SSIM\uparrow$& $CS\uparrow$    \\
    \hline

    IP2P&       0.919 &0.780 &0.268 &43.7277 
&0.094 &0.789 
& 0.669 &  0.274 \\
    \hline
    
 InsDiff&      0.728 &0.648 &0.258 &43.7349 
&0.069 &0.674 
& 0.626 & 0.260 \\
    \hline

 HIVE&      0.847 &0.687 &0.258 &43.7264 
&0.086 &0.789 
& 0.664 & 0.266 \\
    \hline

 MGIE&       0.896 &0.769 &0.258 &43.7265 
&0.070 &0.764 
& 0.671 & 0.257 \\
    \hline
    
 MagicBrush&      0.847 &0.754 &0.267 &43.7261 
&0.044 &0.728 
& 0.672 & 0.274\\
    \hline
    
  HQ-Edit& 0.789& 0.320& 0.251& 43.7266& 0.226& 0.742& 0.388&0.254\\
     \hline
     
  UltraEdit& 0.893& 0.720& 0.268& 43.7410& 0.087& 0.712& 0.589&0.270\\
    \hline
    
    Ours&      \textbf{0.982}&\textbf{0.880}&\textbf{0.273}&\textbf{42.9254}&\textbf{0.032}&\textbf{0.919}& \textbf{0.736}&   \textbf{0.275} \\
        \hline

\end{tabular}

\end{table*}
\begin{table}
    \centering
      \caption{Human preference study results with different methods. $OAE$ represent Object Attribute Editing. $OSP$ represent Object Structure Preservation. $BP$ represent Background Preservation. $OEQ$  represent Overall Editing Quality. }
  \label{tab:t3}
  \begin{tabular}{|c|c|c|c|c|}
        \hline
    Method& $OAE\uparrow$&$OSP\uparrow$&   $BP\uparrow$&$OEQ\uparrow$\\
        \hline
    IP2P&  14.2\%& 7.4\%&  6.2\%&7.1\%\\
        \hline
    InsDiff& 4.1\%& 3.9\%& 4.5\%&1.8\%\\
        \hline
    HIVE& 8.1\%& 7.6\%& 6.3\%&6.5\%\\
        \hline
    MGIE& 5.8\%& 2.1\%& 3.2\%&3.8\%\\
        \hline
    MagicBrush& 10.2\%& 4.3\%& 7.6\%& 7.2\%\\
            \hline
 HQ-Edit& 7.9\%& 2.3\%& 1.2\%&4.3\%\\
         \hline
 UltraEdit& 3.3\%& 5.6\%& 7.6\%&6.4\%\\
        \hline
    Ours& \textbf{46.4\%}& \textbf{66.8\%}& \textbf{63.4\%}& \textbf{62.9\%}\\
        \hline
\end{tabular}

\end{table}

\textbf{Quantitative evaluation.} Given the lack of publicly available datasets for evaluating material and color modification tasks, we developed two specialized datasets designed for quantitative analysis, addressing material and color modification tasks independently.  For the material modification task, we randomly selected 100 subjects and generated 239 source images with StableDiffusion \cite{rombach2022high} checkpoints V1.4 with a random seed. For each source image, we modified the material to all 14 fine-grained materials, resulting in a total of 3,346 pairs of source and target images. For evaluation metrics, we used the DINO score \cite{caron2021emerging} on a gray scale to measure semantic similarity, SSIM \cite{wang2004image} to measure structural similarity, and used the CLIP Score (CS) \cite{radford2021learning} to measure similarity between the target image and its corresponding text description.  For the color modification task, since any object can undergo a color change, we enlarge the subject type to 659 and generate a source image for each subject using the StableDiffusion \cite{rombach2022high} checkpoints V1.4. For each source image, we used the SAM method \cite{kirillov2023segment} to generate the object mask and manually selected the best, resulting in 659 source images with the corresponding subject mask. For each source image, we modified the color to all 43 fine-grained colors, resulting in a total of 28,337 pairs of source and target images. For evaluation metrics, except for using the DINO score,  SSIM, and CLIP score, we used LPIPS \cite{zhang2018unreasonable} in the background area to measure the preservation of background information. L1 loss of HVS of the object area between the target image and the corresponding pure color image to measure how well the color of the object has been changed.  As can be seen in Table \ref{tab:t2}, for color alteration, our method performs best in preserving structure, background information, and the best in changing the object's color compared to other instruct-based methods. For the material modification task, our method performs best in preserving the structure while having the highest text-image similarity after we changed the material of the subject.

\textbf{Human Preference Study.} 
We conduct a human preference study to measure the performance of different methods asking: (1) which method produces the best attributes modification;  (2) which method produces the best preserves the structure of the subject; (3) which method produces the best preserves the background; (4) which method produces the best overall attribute editing quality. We randomly sampled 100 source-target examples and let 20 annotators assess those images. Object attribute editing refers to how well the object's attributes are modified to the target attribute. Object structure preservation measures the extent to which the object's original structure is retained in the edited image. The background preservation indicates how well the background remains intact without distortion. Overall editing quality encompasses the naturalness of the attribute modification, including the overall realism of the image post-editing. The results of the human preference study are listed in Table \ref{tab:t3}.  As can be seen, our method performs the best in all aspects, and the results are consistent with the quantitative experiments. 

%% file: sec/6_conclusion.tex
\section{ Conclusion}
\label{sec:conclusion}
We propose an instruction-based model for object-level color and material modification, enabling precise control over visual attributes while preserving structural and contextual integrity. First, we introduce a training-free Structure Preservation and Attribute Amplification (SPAA) method to generate diverse color/material variations without additional fine-tuning. Using SPAA, we curate the Attribute Dataset, a large-scale resource for object editing, encompassing varied objects, colors, and materials. Training our model on this dataset achieves an optimal balance between accurate attribute editing and preservation of object structure and background. \textbf{Limitations and Future Work.} While SPAA produces plausible variations, cross-attention leakage occasionally causes misaligned edits. Current filtering via LLaVA, Dino, Grounding Dino, and SAM risks excluding a valid target image. Our future work will refine attention mechanisms to constrain focus to target objects, mitigating interference from irrelevant regions.